\documentclass[10pt,twocolumn,letterpaper]{article}

\usepackage{cvpr}              
\usepackage{booktabs,multirow,xcolor,colortbl}
\usepackage{booktabs}
\usepackage{makecell}
\usepackage{graphicx}
\usepackage{colortbl}
\usepackage{tabularx}
\usepackage{array}
\usepackage[table]{xcolor}
\usepackage{adjustbox}
\usepackage{multirow,tabularx,array}
\newcolumntype{Y}{>{\centering\arraybackslash}X}

\definecolor{BestRed}{HTML}{F4B8B5}
\definecolor{SecondLilac}{HTML}{E7E4F5}
\definecolor{OursGray}{HTML}{EFEFEF}

\definecolor{cvprblue}{rgb}{0.21,0.49,0.74}
\usepackage[pagebackref,breaklinks,colorlinks,allcolors=cvprblue]{hyperref}

\def\confName{CVPR}
\def\confYear{2027}

\title{Fuse to Recognize the Unseen: Class-Relative Infrared–Visible Fusion for Open-Vocabulary Semantic Segmentation}

\author{
Zehao Gao\textsuperscript{1}, 
Jie Feng\textsuperscript{2$\dagger$}, 
Wenda Zhao\textsuperscript{2}, 
Junpeng Zhang\textsuperscript{3}, 
Ronghua Shang\textsuperscript{4}, 
Weisheng Dong\textsuperscript{3}\\
\textsuperscript{1} Xidian University, 
\textsuperscript{2} Dalian University of Technology\\
{\tt\small happybirthdayzh@gmail.com, jiefeng0109@163.com}
}

\begin{document}
\maketitle
\begin{abstract}
Visible–infrared image fusion leverages complementary thermal cues to enhance visual perception, particularly under challenging illumination. However, such perceptual complementarity does not readily translate into semantic gains in open-vocabulary semantic segmentation (OVSS), where CLIP-based models are predominantly pretrained on RGB imagery. Directly introducing IR information may instead disrupt RGB–language alignment, turning promising thermal cues into semantic interference that undermines unseen-class recognition. Our pixel-level RGB–IR semantic analysis reveals that class-relevant thermal cues can correct certain RGB misclassifications while preserving reliable RGB predictions, motivating RGB-anchored, class-relative thermal cue selection. Based on this insight, we propose EmergeFuse, an infrared–visible fusion framework for OVSS that selectively converts complementary thermal evidence into unseen-class semantic corrections. Specifically, class-relative cue selection evaluates RGB–IR semantic responses and calibrated thermal evidence relative to RGB predictions, generating class-relative correction requests through competitive selection. Text-contrast image rectification then translates these selected requests into semantic rectification directions through text contrasts, guiding evidence-gated feature updates to realize complementary thermal evidence for unseen-class recognition. Experiments on three benchmarks demonstrate gains of up to 4.755\% in unseen-class mIoU and 15.730\% in strict complementary conversion rate over existing fusion methods, while maintaining competitive fusion quality.
\end{abstract}

\section{Introduction}
\label{sec:intro}

Infrared–visible image fusion integrates visible textures and complementary thermal cues into a unified image for visual interpretation and downstream perception~\cite{1,2,3,4}. By incorporating thermal information, fusion can benefit object detection~\cite{5,6,60} and semantic segmentation~\cite{7,8,9,58,59}, particularly under challenging illumination conditions. To improve downstream performance, task-aware ~\cite{5,7,15} methods employ task-specific losses or integrate downstream features into the fusion process~\cite{6,8,9,16}. More recently, semantic feedback, language conditioning, and preference alignment have been explored to adapt image fusion to diverse perception requirements~\cite{17,18,19,20,21,22,23}.

\begin{figure}[t]
\centering
\includegraphics[width=1\linewidth]{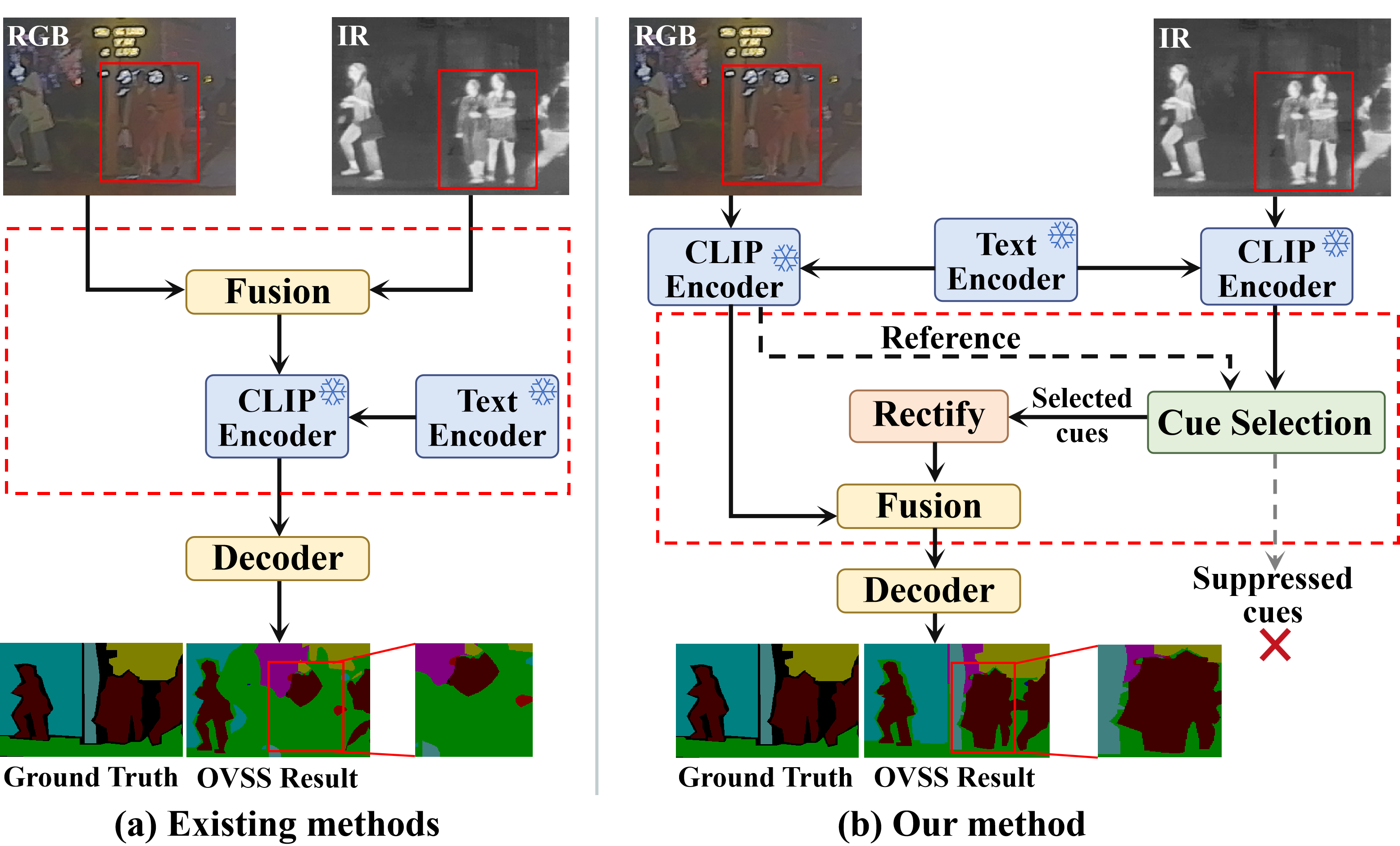}
\caption{Comparison of RGB–IR fusion paradigm OVSS approaches. (a) In a representative fusion-first pipeline, a pedestrian can remain misclassified despite its clear thermal silhouette. (b) EmergeFuse uses each pixel’s RGB prediction as a reference to identify complementary thermal cues before fusion, selectively correcting ambiguous predictions while preserving reliable ones.}
\label{fig:1}
\vspace{-1em}
\end{figure}

However, improving performance on supervised tasks does not necessarily ensure generalization to unseen classes. Open-vocabulary semantic segmentation (OVSS) addresses this challenge by matching visual features with text descriptions via CLIP to recognize categories beyond task-specific training annotations ~\cite{10,11,12,14,56}. This capability motivates the integration of infrared-visible fusion with CLIP-based OVSS to enhance unseen-class recognition, where visual evidence must remain aligned with class semantics beyond supervised classes ~\cite{10,13,24}.

Research on RGB–IR OVSS is still at an early stage, with recent methods such as OpenRSS~\cite{32} and Open-RGBT~\cite{51} exploring thermal information integration through multimodal prompting and vision–language matching. However, since CLIP is predominantly pretrained on RGB imagery, directly incorporating infrared information may disrupt its learned vision–language alignment. These fusion-then-recognition approaches (Fig.~\ref{fig:1}(a)) do not explicitly distinguish whether thermal cues support or refute a queried category relative to the current RGB prediction. Consequently, complementary thermal evidence may fail to correct RGB misclassifications, while misleading cues may disrupt otherwise reliable RGB semantics.

\begin{figure}[t]
\centering
\includegraphics[width=1\linewidth]{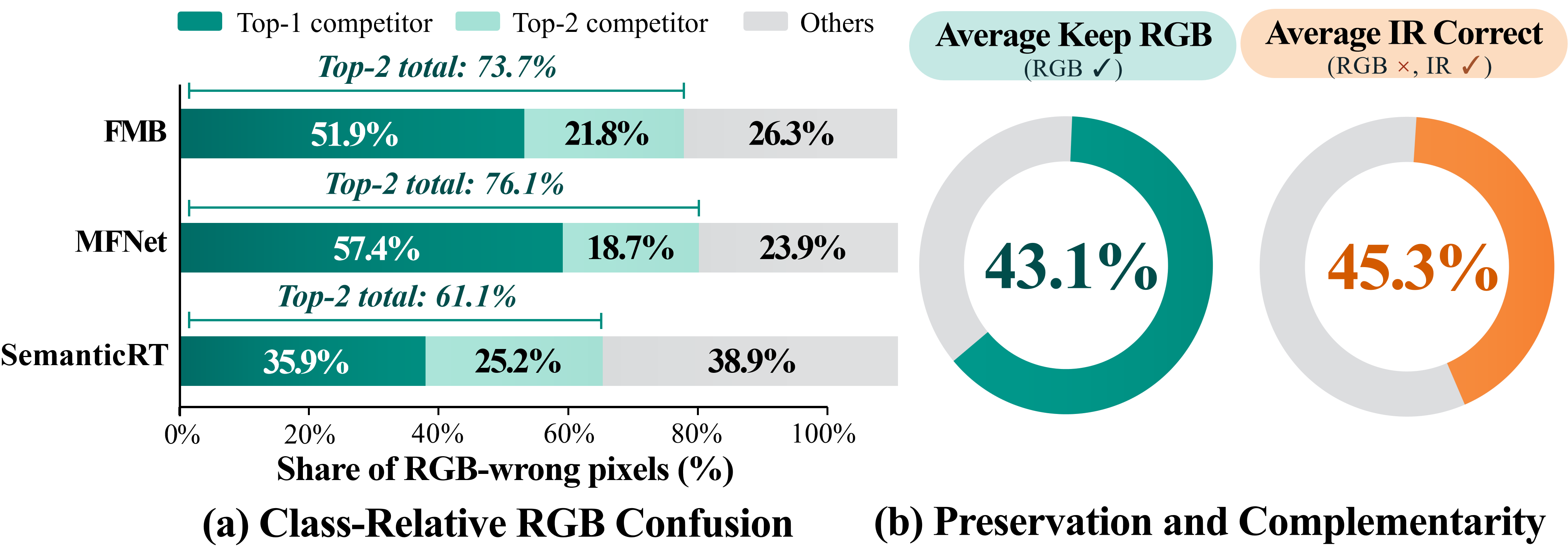}
\caption{Analysis of RGB class confusion and complementary infrared evidence for unseen categories. (a) Class-averaged distributions of incorrect RGB predictions across the top two competing classes and other classes on FMB, MFNet, and SemanticRT. (b) Dataset-averaged proportions of correctly classified RGB pixels (Keep RGB) and RGB-misclassified pixels correctly recognized by infrared (IR Correct).}
\label{fig:2}
\vspace{-0.7em}
\end{figure}

To investigate this issue, we apply the same frozen OVSS reader~\cite{25} separately to RGB and IR and analyze unseen-class predictions across FMB~\cite{8}, MFNet~\cite{64}, and SemanticRT~\cite{65}. As shown in Fig.~\ref{fig:2} (a), under RGB-only input, the top two competing classes account for 73.7\%, 76.1\%, and 61.1\% of RGB misclassifications across three datasets, respectively, revealing concentrated class-dependent confusion patterns. As shown in Fig.~\ref{fig:2} (b), RGB correctly recognizes 43.1\% of unseen-class pixels, while, when using IR alone, 45.3\% of those misclassified by RGB are correctly identified.  These observations motivate RGB-anchored, class-relative thermal cue selection to exploit complementary IR evidence while preserving reliable RGB predictions. See Supplementary Materials for more class analysis.

Guided by these observations, we propose EmergeFuse, which selectively exploits complementary thermal evidence while preserving reliable RGB semantics, as illustrated in Fig.~\ref{fig:1}(b). Class-relative cue selection (CRCS) takes the highest-scoring RGB class at each pixel as a reference and evaluates RGB–IR semantic responses and thermal evidence to identify class-relevant thermal cues that can compensate for RGB misclassifications while suppressing potentially misleading cues. The selected thermal cues are then passed to text-contrast image rectification (TCIR), which derives semantic correction directions from candidate–reference text contrasts to guide their selective integration into multi-scale fusion features, enabling the fused image to express complementary thermal evidence as language-aligned category cues that enhance unseen-class segmentation.

Our main contributions are summarized as follows:
\begin{itemize}

\item We reveal a class-dependent challenge in transferring thermal complementarity to CLIP-based OVSS. Our RGB–IR pixel-level analysis uncovers concentrated class confusion in RGB predictions for unseen categories and substantial IR-only recovery of RGB errors, highlighting the need for class-relative thermal cue selection that balances semantic correction and preservation.

\item We propose EmergeFuse, a class-relative semantic rectification framework that selects complementary thermal evidence to generate targeted rectification requests, then realizes these requests as language-aligned visual cues in fused images for unseen-class recognition without retraining.

\item Experiments on three benchmarks demonstrate gains of up to 4.755\% and 15.730\% in unseen-class mIoU and strict complementary conversion rate, respectively, while maintaining competitive fusion quality.

\end{itemize}

\section{Related Work}
\label{sec:formatting}

\subsection{Task and Language-Guided Image Fusion}

A key challenge in infrared–visible image fusion is to move beyond image-quality-oriented reconstruction and adapt the fusion process to downstream perception requirements. To this end, task-aware methods connect fusion with downstream tasks through task supervision and semantic feature interaction \cite{6,34,35,36,37,38,39}. TDFusion \cite{15} learns fusion objectives from downstream losses via meta-learning, CLDyN \cite{17} adapts fusion through closed-loop semantic compensation, and UniFusion \cite{4} promotes cross-task fusion through modality-consistent representations. DRFusion \cite{62} incorporates task-specific objectives for adaptive image fusion.

Beyond task-driven supervision, language-guided methods exploit textual semantics to provide more flexible control over the fusion process~\cite{40,41,42,50,23}. Text-IF~\cite{18} and GText-IF~\cite{20} leverage textual guidance to adapt fusion to different image degradations, while OmniFuse~\cite{19} extends language-driven control to semantic object enhancement under composite degradations. These advances enable fusion to accommodate diverse task objectives and semantic requirements. However, such adaptability does not necessarily translate into the ability to recognize categories unseen during task-specific training.

\begin{figure*}[h]
\centering
\includegraphics[width=0.9\linewidth]{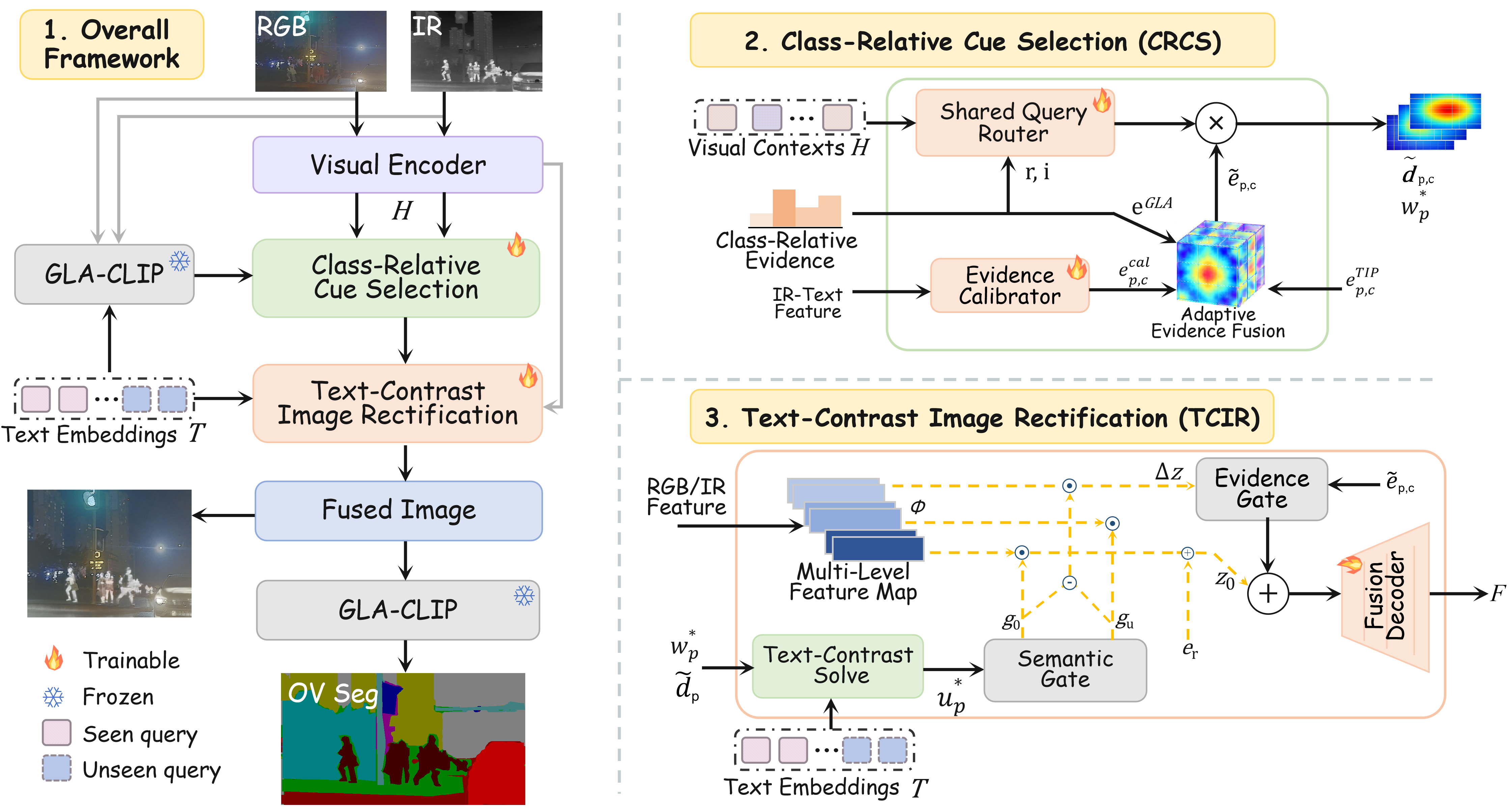}
\caption{Overview of EmergeFuse. CRCS identifies complementary thermal evidence and produces class-relative correction requests, while TCIR translates them into semantic corrections through evidence-constrained text-contrast optimization and multi-scale feature fusion. Frozen readers provide semantic guidance during training and open-vocabulary prediction at inference.}
\label{fig:3}
\vspace{-1.0em}
\end{figure*}

\subsection{Open-Vocabulary Semantic Segmentation}

Open-vocabulary semantic segmentation (OVSS) aims to assign pixel-level labels from text-defined categories, including classes unseen during task-specific training~\cite{43,45,47,48,30,31}. This generalization capability largely relies on CLIP's pretrained vision–language alignment. To improve dense prediction, CAT-Seg~\cite{13} models image–text matching costs, while SED~\cite{26,27} combines hierarchical encoding, gradual cost decoding, and category rejection. Training-free approaches, including SCLIP~\cite{28}, ProxyCLIP~\cite{29}, and GLA-CLIP~\cite{25}, further enhance spatial matching and contextual interactions. Despite their architectural differences, these methods largely build upon visual representations pretrained on RGB imagery, posing challenges for incorporating thermal information while maintaining reliable vision–language alignment.

To extend open-vocabulary recognition beyond visible imagery, multimodal OVSS methods explore incorporating complementary information from additional sensing modalities~\cite{33,52,53,55,57}. OpenRSS~\cite{32} introduces thermal prompts into SAM, while Open-RGBT~\cite{51} combines text and visual prompts for object localization and box-guided mask generation on fused RGB–IR images. These approaches extend the visual evidence available for open-vocabulary recognition, but incorporating thermal information does not necessarily ensure its effective semantic utilization. In contrast, EmergeFuse evaluates thermal evidence relative to RGB-predicted classes and selectively realizes complementary semantic corrections in fused image.

\section{Method}

\subsection{Problem Formulation and Overview}
\label{sec:overview}

RGB–IR open-vocabulary semantic segmentation (OVSS) aims to enhance unseen-class recognition by exploiting complementary visible and infrared information. Given registered RGB and IR images \(R\in[0,1]^{H\times W\times3}\) and \(I\in[0,1]^{H\times W}\), and a query vocabulary \(\mathcal V\), let \(T_{\mathcal V}=\{\mathbf t_c\}_{c\in\mathcal V}\) denote the corresponding embeddings from a frozen CLIP text encoder. Under an image-fusion paradigm, the objective is to generate a fused image \(F\) that supports open-vocabulary recognition through a frozen reader:
\vspace{-1mm}
\begin{equation}
\begin{aligned}
F &= \mathcal F(R,I,T_{\mathcal V}),\\
\hat y_p &= \arg\max_{c\in\mathcal V}
\mathcal R(F,T_{\mathcal V})_{p,c},
\end{aligned}
\label{eq:overview}
\vspace{-1mm}
\end{equation}
where \(\mathcal F\) and \(\mathcal R\) denote the trainable fusion network and frozen OVSS reader, respectively, and \(\hat y_p\) is the predicted category at pixel \(p\). 

To this end, we propose EmergeFuse, as illustrated in Fig.~\ref{fig:3}. A visual encoder extracts multi-scale RGB–IR features, while the frozen OVSS reader provides pixel-level semantic responses for both modalities. These responses, together with calibrated thermal evidence, are fed into Class-Relative Cue Selection (CRCS) to identify complementary thermal cues relative to current RGB predictions and generate class-relative correction requests (Sec.~\ref{sec:CRCS}). The requests are then passed to Text-Contrast Image Rectification (TCIR), where text-contrast semantic directions guide evidence-gated multi-scale feature fusion through source-anchored counterfactual updates (Sec.~\ref{sec:tcir}). Finally, the updated features are decoded into a fused image for segmentation by the same frozen reader to enhance unseen-class recognition while preserving reliable RGB semantics.

\subsection{Class-Relative Cue Selection}
\label{sec:CRCS}

CRCS selects complementary thermal evidence relative to RGB predictions and selects class-relative correction requests. Let $r_{p,c}$ and $i_{p,c}$ denote the semantic scores of category $c\in\mathcal V$ at pixel $p$, obtained by applying the same frozen OVSS reader to RGB image $R$ and infrared image $I$, respectively. Taking the highest-scoring RGB class $a_p=\arg\max_{c\in\mathcal V}r_{p,c}$ as the reference, we estimate the relative thermal advantage of candidate class $c$:
\begin{equation}
e^{\mathrm{GLA}}_{p,c}
=\operatorname{clip}\left(
\frac{(i_{p,c}-i_{p,a_p})-(r_{p,c}-r_{p,a_p})}{\tau_e},
0,1\right),
\label{eq:gla_evidence}
\end{equation}
where $e^{\mathrm{GLA}}_{p,c}$ measures RGB-anchored complementary thermal evidence and $\tau_e$ controls its scale. An eight-dimensional source descriptor \(A_{p,c}\in\mathbb R^8\) is constructed by concatenating source scores, anchor-relative responses, thermal advantage, and cross-modal agreement. The text relation $s^{\mathrm{rel}}_{p,c}=\langle\hat{\mathbf t}_c,\hat{\mathbf t}_{a_p}\rangle$ measures candidate-reference cosine similarity, where $\hat{\mathbf t}$ denotes the $\ell_2$-normalized text embedding.

Since frozen readers may overlook complementary thermal cues, we introduce two additional evidence sources. First, TIP-based evidence $e^{\mathrm{TIP}}_{p,c}$ captures text-aligned TIP-on/off context differences using the thermal information prompt (TIP) module of OpenRSS~\cite{32} retrained with seen-class supervision only. Second, an Evidence Calibrator projects infrared features into the frozen text space to obtain class-relative evidence $e^{\mathrm{cal}}_{p,c}$, supervised by seen-class cross-entropy. The three evidence sources are adaptively fused as
\vspace{-1mm}
\begin{equation}
\begin{aligned}
\boldsymbol{\alpha}_{p,c}
&=\operatorname{Softmax}
\big(g_\alpha(A_{p,c},H_p,s^{\mathrm{rel}}_{p,c})\big),\\
\widetilde e_{p,c}
&=\sum_{k\in\{\mathrm{GLA,TIP,cal}\}}
\alpha^k_{p,c}e^k_{p,c},
\end{aligned}
\label{eq:mixed_evidence}
\end{equation}
where $g_\alpha$ predicts three normalized evidence weights from source descriptors, RGB-IR visual context, and text relations. The resulting $\widetilde e_{p,c}$ provides aggregated complementary evidence for subsequent class-relative cue selection.


Finally, a class-shared router $h_\theta$ generates preliminary signed correction requests:
\begin{equation}
\tilde d_{p,c}
=4\widetilde e_{p,c}
\tanh\big(h_\theta(A_{p,c},H_p,
s^{\mathrm{rel}}_{p,c},\widetilde e_{p,c})\big).
\end{equation}
To suppress unreliable cues, CRCS performs sparse competitive selection over candidate classes and a no-rectification option:
\vspace{-1mm}
\begin{equation}
\begin{aligned}
w_p^\star
=\underset{w_p\in\Delta_p}{\arg\max}
\Bigg[
\sum_{c\neq a_p}w_{p,c}|\tilde d_{p,c}|
+w_{p,\varnothing}b_p^{\mathrm{keep}}
-\frac{\tau_r}{2}\|\boldsymbol{w}_p\|_2^2
\Bigg],
\end{aligned}
\label{eq:query_request}
\end{equation}
where $\Delta_p$ denotes the probability simplex over candidate classes and the no-rectification option, and $\tau_r>0$ controls selection sparsity. The preservation score $b_p^{\mathrm{keep}}=1-|\mathcal V|^{-1}\sum_{c\in\mathcal V}\widetilde e_{p,c}$ favors no rectification when complementary evidence is weak. The resulting class-selection weights $w_{p,c}^\star$ and no-rectification weight $w_{p,\varnothing}^\star$, together with $\tilde d_{p,c}$, guide subsequent text-contrast optimization. Eq.~\eqref{eq:query_request} is efficiently solved via batched simplex projection.

\begin{table*}[t]
\centering
\caption{Comparison of open-vocabulary semantic segmentation on FMB, MFNet, and SemanticRT (\%). All conventional fusion methods and EmergeFuse are evaluated using the same frozen OVSS reader. OpenRSS is included as a cross-framework reference using its native segmentation pipeline. Best and second-best results are highlighted in light red and lavender, respectively.}
\label{tab:1}

\vspace{-0.6mm}
\setlength{\tabcolsep}{2.0pt}
\renewcommand{\arraystretch}{1.05}
\small

\begin{tabularx}{\textwidth}{
l!{\vrule width 0.35pt}
*{4}{Y}!{\vrule width 0.35pt}
*{4}{Y}!{\vrule width 0.35pt}
*{4}{Y}
}
\toprule

\multirow{2}{*}[-5pt]{Method}
& \multicolumn{4}{c!{\vrule width 0.35pt}}{FMB}
& \multicolumn{4}{c!{\vrule width 0.35pt}}{MFNet}
& \multicolumn{4}{c}{SemanticRT} \\

\cmidrule(lr){2-5}
\cmidrule(lr){6-9}
\cmidrule(lr){10-13}

& \shortstack{Seen\\mIoU $\uparrow$}
& \shortstack{Unseen\\mIoU $\uparrow$}
& \raisebox{0.5\baselineskip}[0pt][0pt]{hIoU $\uparrow$}
& \shortstack{Overall\\mIoU $\uparrow$}

& \shortstack{Seen\\mIoU $\uparrow$}
& \shortstack{Unseen\\mIoU $\uparrow$}
& \raisebox{0.5\baselineskip}[0pt][0pt]{hIoU $\uparrow$}
& \shortstack{Overall\\mIoU $\uparrow$}

& \shortstack{Seen\\mIoU $\uparrow$}
& \shortstack{Unseen\\mIoU $\uparrow$}
& \raisebox{0.5\baselineskip}[0pt][0pt]{hIoU $\uparrow$}
& \shortstack{Overall\\mIoU $\uparrow$} \\

\midrule

CDDFuse (CVPR'23)\cite{2}
& 43.219 & 25.773 & 32.290 & 37.851
& 31.052 & 21.251 & 25.233 & 27.377
& 34.218 & 19.467 & 24.816 & 29.301 \\

MRFS (CVPR'24)\cite{9}
& 43.451
& \cellcolor{SecondLilac}\underline{30.464}
& 35.817
& 39.455
& 30.546 & 29.446 & 29.986 & 30.133
& 36.704 & 24.852 & 29.637 & 32.753 \\

TDFusion (CVPR'25)\cite{15}
& \cellcolor{SecondLilac}\underline{46.568}
& 28.769 & 35.566 & \cellcolor{SecondLilac}\underline{41.091}
& \cellcolor{SecondLilac}\underline{34.931}
& 32.329
& \cellcolor{SecondLilac}\underline{33.580}
& \cellcolor{SecondLilac}\underline{33.956}
& \cellcolor{SecondLilac}\underline{38.125}
& 22.761 & 28.505 & 33.004 \\

DCEvo (CVPR'25)\cite{34}
& 44.609 & 27.211 & 33.803 & 39.256
& 33.579 & 31.495 & 32.504 & 32.797
& 35.819 & 23.614 & 28.463 & 31.751 \\

DPOFusion (CVPR'26)\cite{23}
& 43.173 & 27.619 & 33.687 & 38.387
& 29.442 & 30.146 & 29.790 & 29.706
& 33.992 & 21.873 & 26.618 & 29.952 \\

UniFusion (CVPR'26)\cite{4}
& 44.922 & 29.917 & 35.915 & 40.305
& 33.485 & 31.334 & 32.373 & 32.678
& 37.466 & 25.781 & 30.544 & 33.571 \\

DECC (CVPR'26)\cite{66}
& 45.010 & 28.603 & 34.978 & 39.962
& 28.251 & 32.399 & 30.183 & 29.806
& 34.710 & 26.342 & 29.953 & 31.921 \\

DRFusion (CVPR'26)\cite{62}
& 43.747 & 25.316 & 32.072 & 38.076
& 30.207 & 28.184 & 29.160 & 29.449
& 33.276 & 20.719 & 25.537 & 29.090 \\

CLDyN (CVPR'26)\cite{17}
& 41.614 & 27.531 & 33.138 & 37.280
& 31.064
& \cellcolor{SecondLilac}\underline{33.094}
& 32.047 & 31.825
& 32.591
& \cellcolor{SecondLilac}\underline{27.148}
& 29.622 & 30.777 \\

GText-IF (TPAMI'26)\cite{20}
& 44.902 & 30.042 & \cellcolor{SecondLilac}\underline{35.999} & 40.330
& 31.907 & 31.740 & 31.823 & 31.844
& 37.912 & 26.881
& \cellcolor{SecondLilac}\underline{31.457}
& \cellcolor{SecondLilac}\underline{34.235} \\

OpenRSS  (ECCV'24)\cite{32}
& 42.800 & 24.009 & 30.762 & 37.018
& 30.907 & 21.226 & 25.168 & 27.277
& 36.189 & 20.076 & 25.825 & 30.818
\\

\midrule

Ours
& \cellcolor{BestRed}\textbf{47.219}
& \cellcolor{BestRed}\textbf{33.772}
& \cellcolor{BestRed}\textbf{39.379}
& \cellcolor{BestRed}\textbf{43.081}

& \cellcolor{BestRed}\textbf{40.121}
& \cellcolor{BestRed}\textbf{37.849}
& \cellcolor{BestRed}\textbf{38.952}
& \cellcolor{BestRed}\textbf{39.269}

& \cellcolor{BestRed}\textbf{41.328}
& \cellcolor{BestRed}\textbf{30.194}
& \cellcolor{BestRed}\textbf{34.894}
& \cellcolor{BestRed}\textbf{37.617} \\

\bottomrule
\end{tabularx}

\vspace{-1.5mm}
\end{table*}

\subsection{Text-Contrast Image Rectification}
\label{sec:tcir}

TCIR translates the correction requests from CRCS into text-guided feature updates, realizing complementary thermal evidence in the fused image rather than directly modifying segmentation predictions.

\noindent\textbf{Evidence-Constrained Text-Contrast Solving.} Given the correction requests and selection weights from CRCS, we construct a text-contrast matrix $\mathbf D_p$ with rows $\hat{\mathbf t}_c-\hat{\mathbf t}_{a_p}$ and a corresponding request vector $\tilde{\mathbf d}_p$, both indexed by $c\neq a_p$. To balance selected semantic corrections with RGB-anchor preservation, we solve
\vspace{-1mm}
\begin{equation}
\mathbf u_p^\star=\arg\min_{\mathbf u}
\left\{
\|\mathbf D_p\mathbf u-\tilde{\mathbf d}_p\|_{\mathbf W_p}^2
+\mathbf u^\top\boldsymbol{\Lambda}_p\mathbf u
\right\},
\label{eq:contrast_solution}
\end{equation}
where $\mathbf W_p=\operatorname{diag}(w_{p,c}^\star)_{c\neq a_p}$ weights the selected class corrections, and $\boldsymbol{\Lambda}_p=\lambda\mathbf I+\mu{w_{p,\varnothing}^\star}\hat{\mathbf t}_{a_p}\hat{\mathbf t}_{a_p}^{\top}$ regularizes correction magnitude and its component along RGB-anchor text direction. Here, $\|\mathbf x\|_{\mathbf W_p}^2=\mathbf x^\top\mathbf W_p\mathbf x$, with $\lambda>0$ and $\mu\geq0$. This class-shared optimization accommodates unseen text queries without class-specific parameters. Eq.~\eqref{eq:contrast_solution} admits a closed-form solution, evaluated in parallel across the feature grid.

\noindent\textbf{Source-Anchored Counterfactual Fusion.} The visual encoder extracts multi-scale RGB and infrared features $\mathbf E_R$ and $\mathbf E_I$. The pixel-wise directions $\mathbf u_p^\star$ form a spatial guidance map $\mathbf u$, aligned with the multi-scale features. To isolate text-induced corrections from source fusion, we compare the responses of a shared semantic gate $\mathbf g$ with and without semantic guidance.

\noindent\textbf{Evidence-Gated Image Rectification.} To regulate the text-induced corrections, an evidence gate $\boldsymbol{\omega}\in[0,1]$ integrates CRCS thermal evidence, visual context, RGB prediction confidence, and text-visual alignment to regulate the feature updates:
\vspace{-1mm}
\begin{equation}
\begin{aligned}
\widetilde{\mathbf Z}
={}&\mathbf E_R+
\varphi(\mathbf E_I-\mathbf E_R)\odot\\
&\big[\mathbf g(\mathbf 0)+
\boldsymbol{\omega}\odot
(\mathbf g(\mathbf u)-\mathbf g(\mathbf 0))\big].
\end{aligned}
\label{eq:counterfactual_fusion}
\vspace{-1mm}
\end{equation}
Here, $\varphi$ transforms cross-modal feature differences, and $\odot$ denotes element-wise multiplication. These operations are applied at each feature scale. The updated multi-scale features are then reconstructed into a fused image $F$ using a decoder $\mathcal D_\psi$ with RGB detail skip connections, realizing complementary thermal evidence as language-aligned visual cues while preserving reliable RGB semantics.

\subsection{Learning and Seen-to-Unseen Transfer}
\label{sec:learning}

\noindent\textbf{Source-Conditioned Semantic Stopping.} For each supervised pixel $p$ with seen-class label $y_p\in\mathcal V_t$, let $s^X_{p,c}$ denote the frozen-reader score for image $X\in\{R,I,F\}$, where $\mathcal V_t$ is the current training vocabulary and $s^R_{p,c}=r_{p,c}$, $s^I_{p,c}=i_{p,c}$ follow Sec.~\ref{sec:CRCS}. We define the predicted class $\hat y_p^X=\arg\max_{c\in\mathcal V_t}s^X_{p,c}$ and semantic margin $m_p^X=s^X_{p,y_p}-\max_{c\in\mathcal V_t\setminus\{y_p\}}s^X_{p,c}$. Using detached source predictions, we define a target margin for semantic preservation and rectification:
\vspace{-1mm}
\begin{equation}
g_p =
\begin{cases}
\rho_{\mathrm{keep}}[m_p^R]_+,
& \hat y_p^R=y_p,\\
\operatorname{clip}(m_p^I,g_{\min},g_{\max}),
& \hat y_p^R\neq y_p,\ \hat y_p^I=y_p,\\
g_0, & \text{otherwise},
\end{cases}
\label{eq:semantic_target}
\end{equation}
where $[x]_+=\max(x,0)$ and $\rho_{\mathrm{keep}}$ controls RGB margin retention. Here, $g_{\min}$ and $g_{\max}$ bound IR-supported targets, while $g_0$ is the fallback target. These conditions encourage the preservation of correct RGB predictions and the correction of IR-supported RGB errors. We employ a class-balanced semantic stopping loss:
\vspace{-0.5em}
\begin{equation}
\mathcal L_{\mathrm{sem}}
=
\frac{1}{|\mathcal C_t^+|}
\sum_{c\in\mathcal C_t^+}
\frac{1}{|\Omega_c|}
\sum_{p\in\Omega_c}
\left[
\frac{g_p-m_p^F}{\tau_m}
\right]_+,
\label{eq:semantic_stopping}
\vspace{-0.5em}
\end{equation}
where $\Omega_c$ contains supervised pixels of class $c$, $\mathcal C_t^+$ denotes the seen classes present in the current episode, and $\tau_m>0$ controls the loss scale. Once the target margin is satisfied, the semantic penalty vanishes; otherwise, gradients propagate through the frozen reader to optimize the fusion network. During training, we instantiate the query vocabulary $\mathcal{V}$ in Sec.~3.2 as the current episode vocabulary $\mathcal{V}_t$, such that the RGB reference class $a_p$ coincides with $\hat{y}_p^R$.

\noindent\textbf{Seen-to-Unseen Transfer.} Pixel-level supervision is restricted to seen classes, including cross-entropy calibration of the shared infrared–text projection. To improve transferability, we adopt episodic vocabulary variation and prompt consistency to reduce dependence on specific category sets and text prompts. The overall objective combines semantic stopping, evidence calibration, and prompt consistency with SSIM, edge, and color constraints to preserve image fidelity. At inference, CRCS and TCIR apply the same category-shared selection and rectification mechanisms to unseen text queries without retraining. Detailed loss formulations, inference settings, computational complexity, FLOPs,and training protocols are provided in Supplementary Materials.

\section{Experiments}

\begin{table*}[t]
\centering
\caption{Semantic preservation and strict complementary
conversion on FMB, MFNet, and SemanticRT (\%).}
\label{tab:2}

\vspace{-0.6mm}
\setlength{\tabcolsep}{1.85pt}
\renewcommand{\arraystretch}{1.05}
\small

\begin{tabularx}{\textwidth}{
l!{\vrule width 0.35pt}
*{4}{Y}!{\vrule width 0.35pt}
*{4}{Y}!{\vrule width 0.35pt}
*{4}{Y}
}
\toprule

\multirow{2}{*}[-5pt]{Method}
& \multicolumn{4}{c!{\vrule width 0.35pt}}{FMB}
& \multicolumn{4}{c!{\vrule width 0.35pt}}{MFNet}
& \multicolumn{4}{c}{SemanticRT} \\

\cmidrule(lr){2-5}
\cmidrule(lr){6-9}
\cmidrule(lr){10-13}

& \shortstack{Seen\\SPR $\uparrow$}
& \shortstack{Unseen\\SPR $\uparrow$}
& \shortstack{Seen\\SCC $\uparrow$}
& \shortstack{Unseen\\SCC $\uparrow$}

& \shortstack{Seen\\SPR $\uparrow$}
& \shortstack{Unseen\\SPR $\uparrow$}
& \shortstack{Seen\\SCC $\uparrow$}
& \shortstack{Unseen\\SCC $\uparrow$}

& \shortstack{Seen\\SPR $\uparrow$}
& \shortstack{Unseen\\SPR $\uparrow$}
& \shortstack{Seen\\SCC $\uparrow$}
& \shortstack{Unseen\\SCC $\uparrow$} \\

\midrule

CDDFuse (CVPR'23)\cite{2}
& 92.707 & 78.177 & \cellcolor{SecondLilac}\underline{55.060} & 74.907
& 87.768 & 82.064 & 36.937 & \cellcolor{SecondLilac}\underline{63.708}
& 89.216 & 77.538 & 41.735 & 60.214 \\

MRFS (CVPR'24)\cite{9}
& 92.031 & 84.621 & 39.897 & 65.131
& 87.148 & 86.254 & 40.991 & 49.135
& 90.378 & 84.312 & 46.529 & 57.836 \\

TDFusion (CVPR'25)\cite{15}
& 93.382 & \cellcolor{SecondLilac}\underline{85.965}
& 41.537 & \cellcolor{SecondLilac}\underline{75.291}
& 91.563 & 90.328 & 50.000 & 43.302
& 92.416 & 86.574 & 49.331 & 65.217 \\

DCEvo (CVPR'25)\cite{34}
& \cellcolor{SecondLilac}\underline{95.869}
& 85.606 & 54.723 & 73.466
& \cellcolor{SecondLilac}\underline{93.483}
& 89.137 & \cellcolor{SecondLilac}\underline{68.018} & 46.921
& \cellcolor{SecondLilac}\underline{94.157}
& 85.904 & \cellcolor{SecondLilac}\underline{62.517} & 63.895 \\

DPOFusion (CVPR'26)\cite{23}
& 93.217 & 81.565 & 49.776 & 70.219
& 89.139 & 86.537 & 47.229 & 45.487
& 90.781 & 80.193 & 45.842 & 58.197 \\

UniFusion (CVPR'26)\cite{4}
& 94.574 & 83.765 & 46.069 & 66.471
& 92.451 & 89.989 & 53.604 & 52.475
& 93.084 & 86.916 & 57.463 & 67.328 \\

DECC (CVPR'26)\cite{66}
& 93.416 & 84.867 & 41.675 & 68.914
& 82.608 & 88.106 & 59.459 & 62.155
& 88.596 & 83.742 & 52.139 & 68.813 \\

DRFusion (CVPR'26)\cite{62}
& 91.500 & 77.093 & 38.621 & 68.857
& 87.324 & 86.336 & 57.207 & 53.003
& 89.637 & 79.426 & 47.280 & 61.504 \\

CLDyN (CVPR'26)\cite{17}
& 90.842 & 82.208 & 37.263 & 71.124
& 88.784 & 87.300 & 55.856 & 47.775
& 90.125 & 85.170 & 50.478
& \cellcolor{SecondLilac}\underline{70.126} \\

GText-IF (TPAMI'26)\cite{20}
& 95.557 & 84.763 & 47.922 & 59.557
& 91.472 & \cellcolor{SecondLilac}\underline{90.446}
& 54.955 & 55.378
& 93.591 & \cellcolor{SecondLilac}\underline{88.429}
& 58.974 & 66.409 \\

OpenRSS (ECCV'24)\cite{32}
& 91.287 & 78.432 & 37.816 & 57.736
& 84.791 & 80.147 & 34.129 & 47.840
& 88.562 & 77.172 & 56.598 & 59.934 \\

\midrule

Ours
& \cellcolor{BestRed}\textbf{98.902}
& \cellcolor{BestRed}\textbf{95.219}
& \cellcolor{BestRed}\textbf{72.027}
& \cellcolor{BestRed}\textbf{89.349}

& \cellcolor{BestRed}\textbf{98.339}
& \cellcolor{BestRed}\textbf{95.541}
& \cellcolor{BestRed}\textbf{74.324}
& \cellcolor{BestRed}\textbf{79.438}

& \cellcolor{BestRed}\textbf{97.823}
& \cellcolor{BestRed}\textbf{94.168}
& \cellcolor{BestRed}\textbf{76.537}
& \cellcolor{BestRed}\textbf{82.946} \\

\bottomrule
\end{tabularx}
\vspace{-1.5mm}
\end{table*}

\subsection{Experimental Settings}
\label{sec:experimental_settings}

\noindent\textbf{Datasets and Protocol.} We conduct experiments on FMB~\cite{8}, MFNet~\cite{64}, and SemanticRT~\cite{65} under a seen-to-unseen evaluation protocol. The unseen classes are traffic sign, person, truck, and motorcycle on FMB (bicycle has no ground-truth pixels in the test split); person, color cone, and bump on MFNet; and bicyclist, tricycle, traffic light, and person on SemanticRT. Pixel-level semantic supervision is restricted to seen categories, and unseen categories are evaluated without retraining. All fusion methods are evaluated on the same test sets using consistent text queries and a frozen open-vocabulary reader. OpenRSS~\cite{32} is evaluated using its native segmentation pipeline and included as a separate cross-framework reference. GLA-CLIP~\cite{25} provides semantic supervision during fusion training.

\noindent\textbf{Evaluation Metrics.} We report seen, unseen, and overall mIoU, together with the harmonic mean (hIoU) of seen and unseen mIoU. To evaluate semantic preservation and complementary conversion, we measure semantic preservation rate (SPR) and strict complementary conversion rate (SCC) for both seen and unseen classes. SCC measures the recovery rate on a fixed set of pixels misclassified by RGB but correctly recognized by IR. Fusion quality is assessed using VIF, Qabf, SSIM, and SF. Detailed metric definitions are provided in Supplementary Materials.

\subsection{Open-Vocabulary Performance and Semantic Conversion}
\label{sec:quantitative_results}

Table~\ref{tab:1} compares OVSS performance on FMB, MFNet, and SemanticRT. As few existing methods directly address open-vocabulary-aware infrared–visible fusion, we primarily compare against representative fusion methods. Open-RGBT~\cite{51} is excluded from the quantitative comparison because its official implementation is unavailable. EmergeFuse achieves the highest hIoU. Its unseen mIoU reaches 33.772\%, 37.849\%, and 30.194\%, outperforming the strongest conventional fusion baselines by 3.308, 4.755, and 3.046 percentage points, respectively. These improvements highlight the effectiveness of CRCS in selecting complementary thermal evidence and TCIR in rectifying fused-image semantics for unseen-class recognition.

\begin{table}[t]
\centering
\caption{Ablation studies of EmergeFuse on FMB (\%).
``--'' denotes a component not instantiated in the corresponding variant.}
\label{tab:3}
\vspace{-0.6mm}
\small
\setlength{\tabcolsep}{2pt}
\renewcommand{\arraystretch}{1.12}

\resizebox{\columnwidth}{!}{%
\begin{tabular}{
@{}l
!{\vrule width 0.35pt}
cccc
!{\vrule width 0.35pt}
ccc@{}
}
\toprule

\makecell[l]{Variant}
& \makecell[c]{CRCS}
& \makecell[c]{TCIR}
& \makecell[c]{Evidence\\Calibrator}
& \makecell[c]{Semantic\\Gate}
& \makecell[c]{Unseen\\mIoU $\uparrow$}
& \makecell[c]{hIoU $\uparrow$}
& \makecell[c]{Unseen\\SCC $\uparrow$} \\

\midrule

Basic
& $\times$ & $\times$ & $\times$ & $\times$
& 26.384 & 32.294 & 58.573 \\

CRCS-only
& $\checkmark$ & $\times$ & -- & --
& 30.868 & 36.579 & 71.270 \\

TCIR-only
& $\times$ & $\checkmark$ & -- & --
& 31.205 & 36.837 & 62.722 \\

w/o Calib
& $\checkmark$ & $\checkmark$ & $\times$ & $\checkmark$
& 31.853 & 37.723 & 83.960 \\

w/o Gate
& $\checkmark$ & $\checkmark$ & $\checkmark$ & $\times$
& 32.120 & 37.844 & 82.804 \\

\midrule
\rowcolor{OursGray}
Full
& $\checkmark$ & $\checkmark$ & $\checkmark$ & $\checkmark$
& \textbf{33.772} & \textbf{39.379}
& \textbf{89.349} \\

\bottomrule
\end{tabular}%
}
\vspace{-1em}
\end{table}

Table~\ref{tab:2} evaluates semantic preservation and complementary conversion on three datasets. EmergeFuse achieves SCC Unseen of 89.349\%, 79.438\%, and 82.946\%, exceeding the strongest competitors by 14.058\%, 15.730\%, and 12.820\%, respectively. Compared with OpenRSS, EmergeFuse further improves SCC Unseen by 31.613\%, 31.598\%, and 23.012\%. These results highlight the effectiveness of CRCS and TCIR in converting complementary thermal evidence into recognizable unseen-class semantics while preserving reliable RGB predictions.

\begin{table}[t]
\centering
\caption{Controlled comparison of semantic supervision
and text-conditioned fusion on FMB (\%).}
\label{tab:6}
\vspace{-0.6mm}
\small
\setlength{\tabcolsep}{2pt}
\renewcommand{\arraystretch}{1.08}

\begin{tabular}{@{}lccccc@{}}
\toprule

\makecell[l]{Method}
& \makecell[c]{Seen\\mIoU $\uparrow$}
& \makecell[c]{Unseen\\mIoU $\uparrow$}
& \makecell[c]{hIoU $\uparrow$}
& \makecell[c]{Unseen\\SCC $\uparrow$}
& \makecell[c]{Unseen\\SPR $\uparrow$} \\

\midrule

Basic
& 41.617 & 26.384 & 32.294 & 58.573 & 84.729 \\

Basic+Sem
& 45.127 & 30.836 & 36.637 & 78.146 & 90.853 \\

Basic+Sem+FiLM
& 46.039 & 32.118 & 37.839 & 82.327 & 92.741 \\

\midrule

\rowcolor{OursGray}
Ours
& \textbf{47.219}
& \textbf{33.772}
& \textbf{39.379}
& \textbf{89.349}
& \textbf{95.219} \\

\bottomrule
\end{tabular}
\vspace{-1em}
\end{table}

\begin{figure}[h]
\centering
\includegraphics[width=1\linewidth]{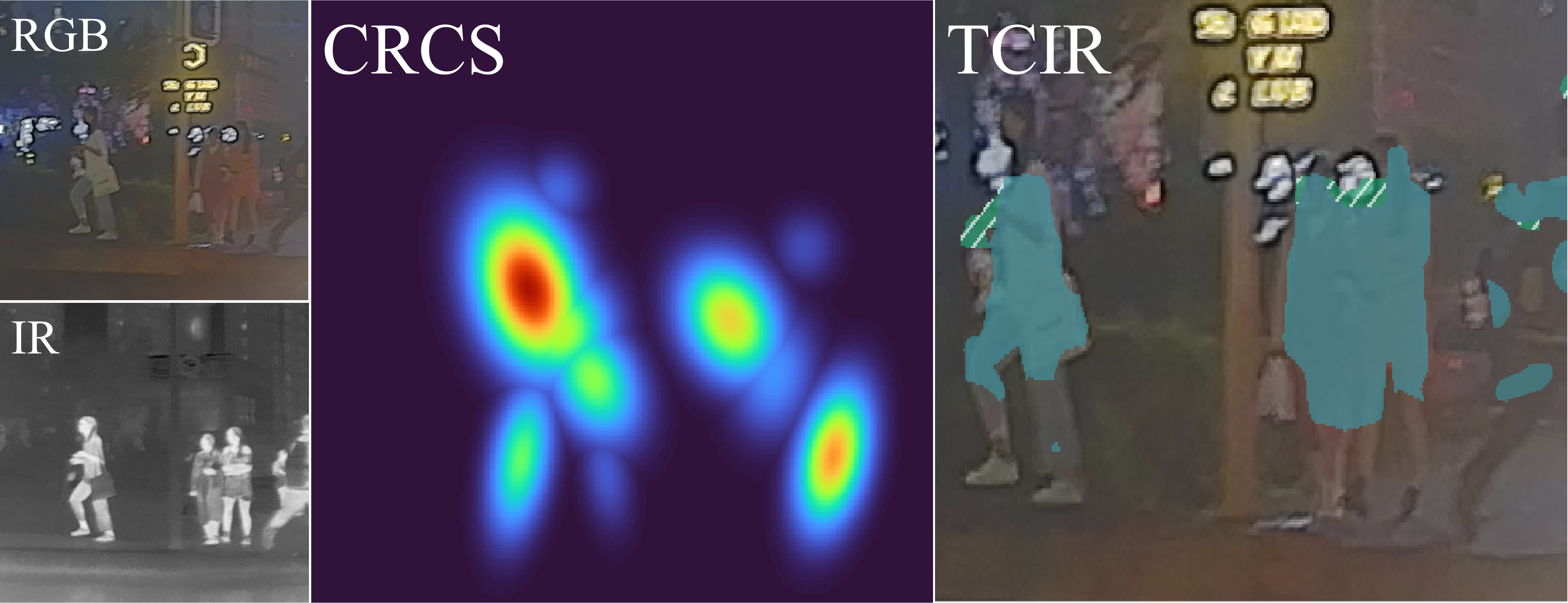}
\caption{Mechanism visualization for the unseen person category. From left to right: RGB/IR inputs, CRCS response map, and TCIR-guided semantic outcomes (cyan: corrected RGB errors; green: preserved RGB-correct pixels).}
\label{fig:mechanism_vis}
\vspace{-1em}
\end{figure}

\begin{table*}[t]
\centering
\caption{Quantitative comparison of infrared-visible image fusion quality on FMB, MFNet, and SemanticRT. SSIM is computed with the RGB input as the reference. Additional metrics (such as $\Delta$Margin, EN, SD, and AG) are provided in Supplementary Materials.}
\label{tab:4}

\vspace{-0.6mm}
\small
\setlength{\tabcolsep}{1.8pt}
\renewcommand{\arraystretch}{1.05}

\begin{tabularx}{\textwidth}{
l!{\vrule width 0.35pt}
*{4}{Y}
!{\vrule width 0.35pt}
*{4}{Y}
!{\vrule width 0.35pt}
*{4}{Y}
}
\toprule

\multirow{2}{*}{Method}
& \multicolumn{4}{c!{\vrule width 0.35pt}}{FMB}
& \multicolumn{4}{c!{\vrule width 0.35pt}}{MFNet}
& \multicolumn{4}{c}{SemanticRT} \\

\cmidrule(lr){2-5}
\cmidrule(lr){6-9}
\cmidrule(lr){10-13}

& VIF $\uparrow$
& Qabf $\uparrow$
& \shortstack{SSIM $\uparrow$}
& SF $\uparrow$
& VIF $\uparrow$
& Qabf $\uparrow$
& \shortstack{SSIM $\uparrow$}
& SF $\uparrow$
& VIF $\uparrow$
& Qabf $\uparrow$
& \shortstack{SSIM $\uparrow$}
& SF $\uparrow$ \\

\midrule

CDDFuse (CVPR'23)\cite{2}
& 0.875 & 0.688 & 0.930
& \cellcolor{SecondLilac}\underline{14.575}
& \cellcolor{SecondLilac}\underline{1.015}
& 0.699 & 0.700
& 11.422
& 0.886 & 0.682 & 0.902 & 14.914 \\

MRFS (CVPR'24)\cite{9}
& 0.467 & 0.189 & 0.829 & 5.181
& 0.567 & 0.226 & 0.517 & 4.794
& 0.493 & 0.230 & 0.813 & 5.376 \\

TDFusion (CVPR'25)\cite{15}
& 0.840 & 0.678 & 0.919 & 14.250
& 1.012
& \cellcolor{SecondLilac}\underline{0.710}
& \cellcolor{SecondLilac}\underline{0.795}
& 11.696
& 0.901 & 0.690 & 0.915 & 14.203 \\

DCEvo (CVPR'25)\cite{34}
& 0.888
& \cellcolor{SecondLilac}\underline{0.699}
& 0.932 & 13.751
& 0.909 & 0.676 & 0.727 & 11.920
& 0.922
& \cellcolor{SecondLilac}\underline{0.706}
& 0.925 & 14.631 \\

DPOFusion (CVPR'26)\cite{23}
& 0.797 & 0.628 & 0.914 & 13.172
& 0.767 & 0.582 & 0.694 & 9.298
& 0.780 & 0.615 & 0.887 & 12.845 \\

UniFusion (CVPR'26)\cite{4}
& 0.718 & 0.635 & 0.933 & 14.404
& 0.804 & 0.624 & 0.661 & 11.782
& 0.764 & 0.648 & 0.909 & 13.624 \\

DECC (CVPR'26)\cite{66}
& 0.801 & 0.668 & 0.920 & 13.860
& 0.361 & 0.568 & 0.112
& \cellcolor{BestRed}\textbf{12.982}
& 0.806 & 0.671 & 0.838
& \cellcolor{BestRed}\textbf{16.302} \\

DRFusion (CVPR'26)\cite{62}
& \cellcolor{SecondLilac}\underline{0.918}
& 0.670 & 0.902 & 13.103
& 0.892 & 0.615 & 0.635 & 11.293
& \cellcolor{BestRed}\textbf{0.952}
& 0.666 & 0.899 & 13.518 \\

CLDyN (CVPR'26)\cite{17}
& 0.364 & 0.602 & 0.599 & 10.457
& 0.460 & 0.630 & 0.383 & 9.299
& 0.431 & 0.587 & 0.636 & 10.743 \\

GText-IF (TPAMI'26)\cite{20}
& 0.814 & 0.674
& \cellcolor{SecondLilac}\underline{0.944}
& 12.948
& 0.880 & 0.614 & 0.658 & 10.502
& 0.843 & 0.688
& \cellcolor{SecondLilac}\underline{0.936}
& 13.752 \\

\midrule

Ours
& \cellcolor{BestRed}\textbf{0.933}
& \cellcolor{BestRed}\textbf{0.736}
& \cellcolor{BestRed}\textbf{0.991}
& \cellcolor{BestRed}\textbf{18.817}
& \cellcolor{BestRed}\textbf{1.048}
& \cellcolor{BestRed}\textbf{0.736}
& \cellcolor{BestRed}\textbf{0.957}
& \cellcolor{SecondLilac}\underline{11.941}
& \cellcolor{SecondLilac}\underline{0.927}
& \cellcolor{BestRed}\textbf{0.747}
& \cellcolor{BestRed}\textbf{0.958}
& \cellcolor{SecondLilac}\underline{15.486} \\

\bottomrule
\end{tabularx}
\vspace{-2mm}
\end{table*}

\begin{table}[t]
\centering
\caption{Cross-reader generalization of EmergeFuse on FMB (\%). RGB-U denotes unseen mIoU using RGB-only input. $\Delta$U mIoU denotes the unseen-class mIoU gain over RGB-U.  }
\label{tab:5}

\vspace{-0.6mm}
\small
\setlength{\tabcolsep}{2.0pt}
\renewcommand{\arraystretch}{1.05}

\begin{tabularx}{\columnwidth}{
>{\raggedright\arraybackslash}m{23mm}
!{\vrule width 0.35pt}
*{4}{Y}
}

\toprule
\smash{\raisebox{0.35\baselineskip}{Reader}}
& \shortstack{RGB-U\\mIoU $\uparrow$}
& \shortstack{Seen\\mIoU $\uparrow$}
& \shortstack{Unseen\\mIoU $\uparrow$}
& \shortstack{$\Delta$U\\mIoU $\uparrow$} \\

\midrule

SCLIP\cite{28}
& 24.231 & 27.501 & 29.253 & +5.022 \\

NACLIP\cite{46}
& 27.471 & 44.079 & 30.374 & +2.903 \\

ProxyCLIP\cite{29}
& 28.727 & 45.490 & 31.260 & +2.533 \\

\midrule

\rowcolor{OursGray}
GLA-CLIP\cite{25}
& \textbf{29.853}
& \textbf{47.219}
& \textbf{33.772}
& \textbf{+3.919} \\

\bottomrule
\end{tabularx}
\vspace{-1.5mm}
\end{table}

\subsection{Ablation Studies}
\label{sec:ablation}

Table~\ref{tab:3} evaluates EmergeFuse components on FMB. Basic retains the fusion backbone and decoder with reconstruction losses only. CRCS-only replaces TCIR with lightweight evidence-conditioned fusion, while TCIR-only uses fixed source-relative evidence instead of learned routing. Both outperform Basic. The inclusion of calibration and gating yielded improvements of 1.919\% and 1.652\%, respectively, demonstrating the complementary benefits of cue selection and rectification in unseen-class recognition.


\begin{table}[t]
\centering
\caption{Robustness to vocabulary perturbations and
textual variations on FMB (\%).}
\label{tab:vocab_fmb}

\vspace{-0.6mm}
\small
\setlength{\tabcolsep}{1.85pt}
\renewcommand{\arraystretch}{1.05}

\begin{tabularx}{\columnwidth}{
>{\raggedright\arraybackslash}m{23mm}
!{\vrule width 0.35pt}
*{3}{Y}
}

\toprule
\smash{\raisebox{0.35\baselineskip}{Condition}}
& \shortstack{Unseen\\mIoU $\uparrow$}
& \shortstack{Unseen\\SPR $\uparrow$}
& \shortstack{Unseen\\SCC $\uparrow$} \\

\midrule

Random+4
& \cellcolor{SecondLilac}\underline{33.558}
& \cellcolor{SecondLilac}\underline{95.087}
& 88.812 \\

Hard+4
& 33.126 & 94.704 & 87.684 \\

Hard+8
& 32.417 & 94.109 & 85.906 \\

Prompt Shift
& 33.491
& 94.992
& \cellcolor{SecondLilac}\underline{88.938} \\

\midrule
Original
& \cellcolor{BestRed}\textbf{33.772}
& \cellcolor{BestRed}\textbf{95.219}
& \cellcolor{BestRed}\textbf{89.349} \\

\bottomrule
\end{tabularx}
\vspace{-2mm}
\end{table}

Table~\ref{tab:6} presents a controlled comparison on FMB, starting from the Basic variant in Table~\ref{tab:3}. We progressively introduce semantic supervision (+Sem) and text-conditioned feature modulation (+FiLM), with both variants using the same semantic loss as EmergeFuse. While these additions improve unseen-class recognition, EmergeFuse further outperforms Basic+Sem+FiLM by 1.654 points in unseen mIoU and 7.022 points in unseen SCC, while achieving 95.219\% unseen SPR. These gains demonstrate the benefits of CRCS and TCIR beyond conventional semantic supervision and text-conditioned fusion. 

\begin{table}[t]
\centering
\small
\caption{Cross-dataset zero-shot generalization of EmergeFuse.}
\label{tab:cross_dataset}
\vspace{-0.6mm}
\setlength{\tabcolsep}{3pt}
\begin{tabular}{lccc}
\toprule
\multirow{2}{*}{Transfer} & In-Domain & Cross-Data & In-Domain \\
         & Unseen  $\uparrow$  & Unseen $\uparrow$
         & Runner-up  $\uparrow$  \\
\midrule
FMB $\rightarrow$ MFNet
    & 37.849 & 31.157 & 33.094 \\
FMB $\rightarrow$ SemanticRT
    & 30.194 & 28.347 & 27.148 \\
\bottomrule
\end{tabular}
\vspace{-2mm}
\end{table}

\begin{figure*}[h]
\centering
\includegraphics[width=1\linewidth]{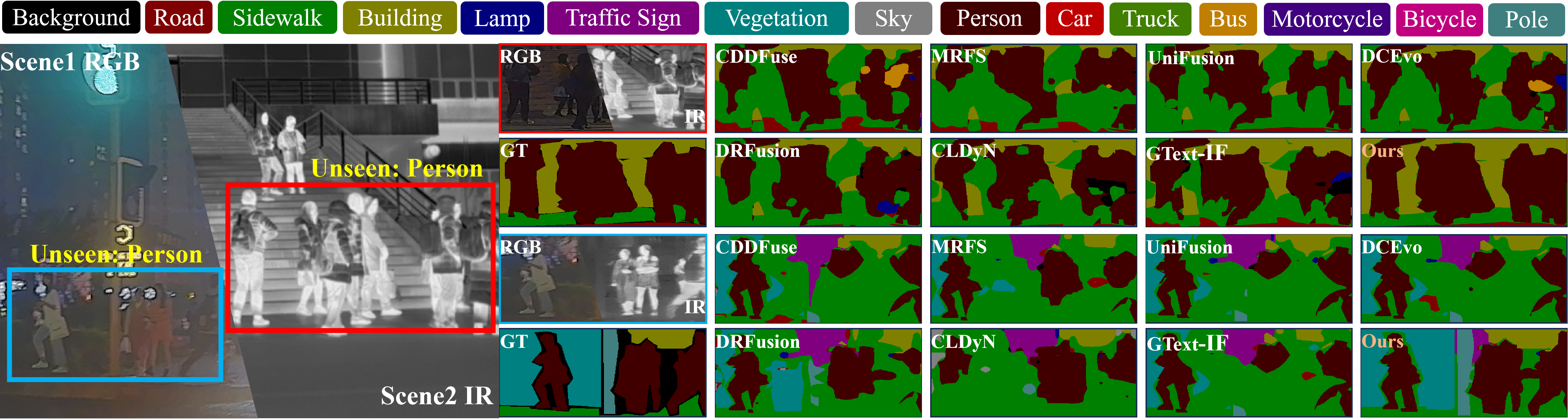}
\caption{Qualitative comparison of open-vocabulary segmentation on FMB. The left panel shows two representative scenes with highlighted unseen person regions, the middle panel presents corresponding RGB and IR crops and ground-truth masks, and the right panel compares segmentation predictions from different methods.}
\label{fig:OVSS_vis}
\end{figure*}

\begin{figure*}[h]
\centering
\includegraphics[width=1\linewidth]{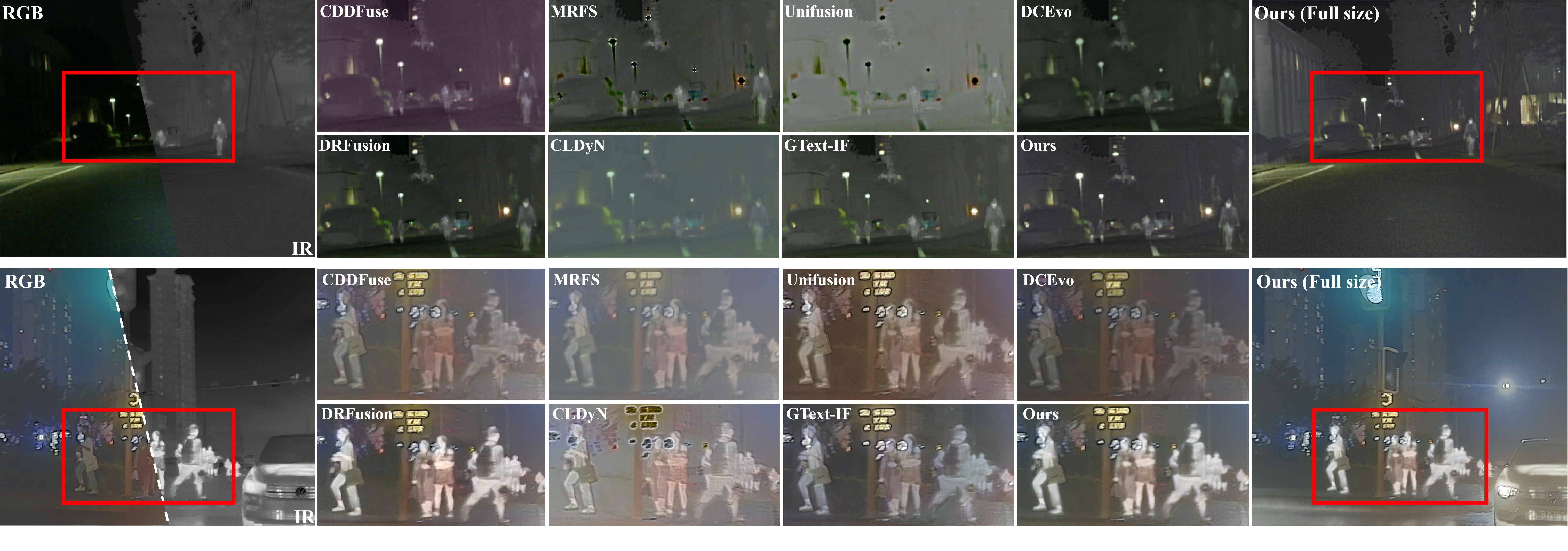}
\vspace{-2.2em}
\caption{Qualitative comparison of infrared–visible image fusion on FMB and MFNet. Two representative nighttime scenes are presented. The left panels show RGB and infrared inputs with highlighted regions, the middle panels compare the corresponding enlarged fusion results, and the right panels display the full-resolution fused images generated by EmergeFuse.}
\label{fig:fusion_vis}
\vspace{-1em}
\end{figure*}

Fig.~\ref{fig:mechanism_vis} illustrates how EmergeFuse selectively corrects RGB errors while preserving reliable predictions. CRCS highlights pedestrian regions with complementary thermal evidence, while TCIR realizes targeted semantic corrections without disrupting RGB-correct regions. This behavior reflects RGB-anchored cue selection and text-contrast rectification, consistent with the improvements in SCC and SPR.

\subsection{Fusion Quality and Generalization}
\label{sec:generalization}

Table~\ref{tab:4} evaluates image fusion quality on FMB, MFNet, and SemanticRT. EmergeFuse achieves strong performance across standard fusion metrics, demonstrating that improved semantic recognition is accompanied by competitive fusion quality. EmergeFuse employs a compact trainable fusion core, with most of its parameters residing in the frozen pretrained context extractor and OVSS reader. The trainable fusion network contains only 0.67M parameters.

\noindent\textbf{Cross-Reader Generalization.} Table~\ref{tab:5} evaluates cross-reader generalization on FMB. Beyond the training reader, SCLIP, NACLIP, and ProxyCLIP achieve unseen mIoU gains of 2.533–5.022 points over their RGB-only baselines. These results demonstrate the transferability of semantic evidence embedded in EmergeFuse outputs.

\noindent\textbf{Cross-Dataset Zero-Shot Generalization.} As shown in Table~\ref{tab:cross_dataset}, EmergeFuse trained solely on FMB achieves 31.157\% and 28.347\% unseen mIoU on MFNet and SemanticRT without target-domain adaptation, respectively. Notably, our model, trained on FMB and transferred directly to SemanticRT in a zero-shot manner (28.347\%), even outperforms CLDyN (27.148\%), the strongest baseline trained on the SemanticRT domain itself. These results demonstrate promising cross-dataset generalization, supporting the transferability of class-relative cue selection.

\noindent\textbf{Vocabulary Robustness.} Table~\ref{tab:vocab_fmb} evaluates vocabulary robustness by adding random (Random+4) or challenging (Hard +4/+8) distractor classes or varying text prompts (prompt shift) to the fusion-stage vocabulary, while keeping the evaluation vocabulary fixed. Under Hard+8, unseen mIoU decreases by only 1.355 points, while unseen SCC remains at 85.906\%. These results demonstrate the robustness of EmergeFuse to vocabulary perturbations. 

\subsection{Qualitative Results}
\label{sec:generalization}

Fig.~\ref{fig:OVSS_vis} presents OVSS comparisons on FMB. In both scenes, several fusion methods produce fragmented unseen-person regions or confuse targets with surrounding classes. In contrast, EmergeFuse generates more complete segmentation masks with improved agreement. These observations complement the quantitative gains in unseen mIoU and SCC, supporting the effectiveness of CRCS and TCIR. 
Fig.~\ref{fig:fusion_vis} presents fusion comparisons on FMB and MFNet. Several existing methods exhibit reduced contrast or weakened structural details in challenging nighttime scenes. In contrast, EmergeFuse preserves recognizable thermal targets while producing visually coherent fused images.

\section{Conclusion}


Effective RGB–IR fusion for OVSS requires selectively translating complementary thermal evidence into language-aligned visual semantics. EmergeFuse achieves this through RGB-anchored guidance and image-level rectification, improving unseen-class recognition while preserving reliable RGB predictions. Experiments on three benchmarks demonstrate improved unseen-class segmentation with competitive fusion quality.

{
    \small
    \bibliographystyle{ieeenat_fullname}
    \bibliography{main}
}



\end{document}